\documentclass[10pt,twocolumn,letterpaper]{article}

\usepackage[pagenumbers]{cvpr} 

\usepackage{graphicx}

\usepackage{amsmath}
\usepackage{amssymb}
\usepackage{booktabs}
\usepackage{amssymb}
\usepackage{pifont}
\newcommand{\cmark}{\ding{51}}%
\newcommand{\xmark}{\ding{55}}%
\usepackage{nccmath}
\usepackage{xcolor, soul}
\sethlcolor{yellow}
\usepackage[pagebackref,breaklinks,colorlinks]{hyperref}

\usepackage[capitalize]{cleveref}
\crefname{section}{Sec.}{Secs.}
\Crefname{section}{Section}{Sections}
\Crefname{table}{Table}{Tables}
\crefname{table}{Tab.}{Tabs.}

\makeatletter
\newcommand{\printfnsymbol}[1]{%
\textsuperscript{\@fnsymbol{#1}}%
}
\makeatother

\begin{document}

	\title{LatentFormer: Multi-Agent Transformer-Based Interaction Modeling and Trajectory Prediction} 
\author{Elmira Amirloo\thanks{equal contribution}\thanks{This work was done while the author was a Huawei employee. }, Amir Rasouli\printfnsymbol{1}, Peter Lakner, Mohsen Rohani, Jun Luo \\
Noah's Ark Lab, Huawei, Toronto, Canada \\
 \\
{\tt\small  $\{$amir.rasouli, peter.lakner, mohsen.rohani, jun.luo1$\}$@huawei.com} 

\and

}
	
	\maketitle
	
    \vspace{-30pt}
	\vspace{-20pt}
\begin{abstract}
Multi-agent trajectory prediction is a fundamental problem in  autonomous driving. The key challenges in prediction are accurately anticipating the behavior of surrounding agents and understanding the scene context. To address these problems, we propose LatentFormer, a transformer-based model for predicting future vehicle trajectories. The proposed method leverages a novel technique for modeling interactions among dynamic objects in the scene. Contrary to many existing approaches which model cross-agent interactions during the observation time, our method additionally exploits the future states of the agents. This is accomplished using a hierarchical attention mechanism where the evolving states of the agents autoregressively control the contributions of past trajectories and scene encodings in the final prediction. Furthermore, we propose a multi-resolution map encoding scheme that relies on a vision transformer module to effectively capture both local and global scene context to guide the generation of more admissible future trajectories. We evaluate the proposed method on the nuScenes benchmark dataset and show that our approach achieves state-of-the-art performance and improves upon trajectory metrics by up to 40\%. We further investigate the contributions of various components of the proposed technique via extensive ablation studies.
\end{abstract}

	\vspace{-10pt}
\section{Introduction}

Predicting road user behavior is one of the fundamental problems in autonomous driving. In this context, prediction often is realized in the form of forecasting the future trajectories of the road users, such as pedestrians and vehicles \cite{Kothari_2021_CVPR,Shafiee_2021_CVPR,Kim_2021_CVPR,Cui_2021_ICCV,Dendorfer_2021_ICCV}. Predictions are then used for planning future actions.  

Trajectory prediction is inherently a challenging task as it requires in-depth understanding of agents' dynamics as well as various environmental factors that are not directly observable. As demonstrated by a large body of work in this domain \cite{Ma_2021_ICCV, Narayanan_2021_CVPR,Park_2020_ECCV,Salzmann_2020_ECCV,Casas_2020_ECCV}, there are two components that play key roles in predicting trajectories: modeling interactions between the agents and understanding environmental constraints (e.g. road structure, drivable areas). The former is important as the behavior of one agent, e.g. obstructing the road, can have a direct impact on the behavior of another agent. The latter gives the system the ability to reason about whether a predicted trajectory is valid, e.g. it is within the drivable area.

In practice, when predicting an agent's behavior, there are often multiple possibilities that one should consider. For instance, when a vehicle is approaching an intersection, turning right or moving straight may be equally valid choices. As a result, when designing a predictive model, it is important to encourage a diversity of predictions while ensuring the predicted futures are valid or, as termed in \cite{Park_2020_ECCV}, admissible. 

\begin{figure}
\centering
\includegraphics[width=1\columnwidth]{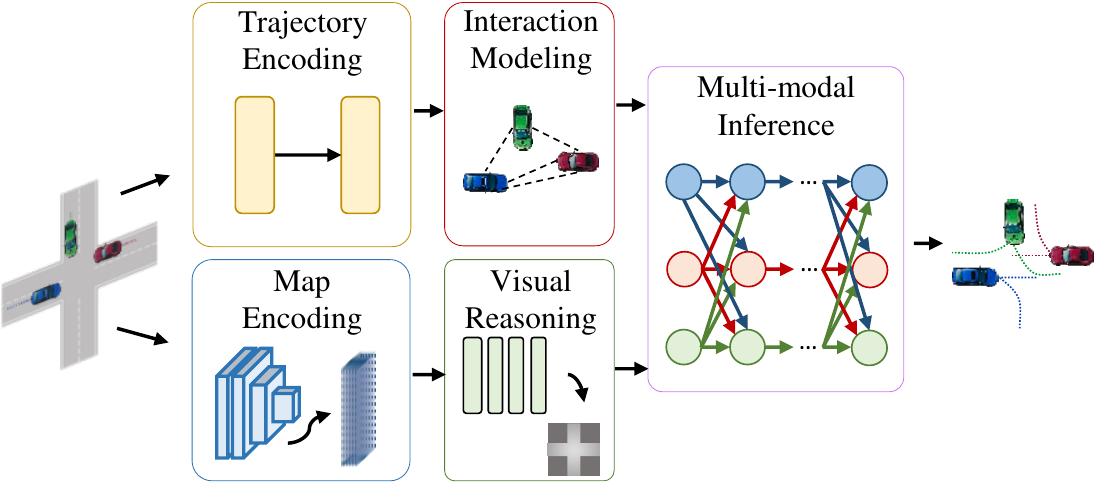}
\caption{An overview of the proposed method, LatentFormer. Our method relies on observed trajectories and map representations to model interactions among agents and between agents and the environment. The inference is done jointly for all agents to produce globally coherent trajectories.}
\vspace{-0.3cm}
\label{first_image}
\end{figure}

In this paper, we address the aforementioned challenges in the context of vehicle trajectory prediction. More specifically, we propose LatentFormer, a novel multi-modal trajectory prediction model which relies on a transformer-based architecture to capture the complex scene context and generate diverse future trajectories as shown in Figure \ref{first_image}. Our method uses a self-attention module that encodes the interactions between agents according to their dynamics. Unlike many existing approaches \cite{Park_2020_ECCV, Chai_2019_CORL, Salzmann_2020_ECCV}, we autoregressively take into account the future states of all the agents in the decoding phase to model how they interact with one another. This allows the model to explicitly account for the expected future interactions of agents and avoid conflicting state predictions.

To encode the scene context, we use a multi-resolution approach where local and global patches are extracted from the map. In order to alleviate location invariance bias of convolutional encoding methods, we use a vision transformer to focus on parts of the map representations that are relevant to each agent. Our proposed prediction module follows a hierarchical attention architecture where the predicted states of the agents serve as a query to update the belief of the system regarding the past observations and the scene context. To encourage multi-modality in the system, the proposed model uses the past trajectories and the map information to infer \textit{``intentions"} of agents. These intentions are then used as discrete modes in a Conditional Variational AutoEncoder (CVAE) framework which enables estimating a multi-modal distribution over trajectories. 

\textbf{Contributions} of our work are as follows: \textbf{1)} We propose a novel transformer-based architecture for trajectory prediction in a multi-agent setting. The proposed model produces diverse multi-modal predictions that effectively capture the dynamic nature of the road users. By evaluating the model on nuScenes benchmark dataset, we show that our method achieves state-of-the-art performance. \textbf{2)} We propose a novel attention-based technique for modeling interactions between the road agents. We conduct ablation studies to demonstrate the contribution of this module to the overall performance. \textbf{3)} We propose a multi-resolution map encoding scheme in which both local and global information is extracted from the scenes. Using a vision transformer method, the map representation is linked to the state of the agents to model their interactions with the scene. We evaluate the contribution of our map encoding technique via ablation studies. \textbf{4)} We study the possibility of using the model in a non-autoregressive mode to enhance the computational efficiency of the proposed model in inference mode. We investigate the effectiveness of each approach via empirical studies.

	\section{Related Works}

\subsection{Trajectory prediction methods}
Trajectory prediction is of vital importance to many artificial intelligent applications. There is a large body of works on this topic designed to predict behavior of pedestrians \cite{Kothari_2021_CVPR,Shafiee_2021_CVPR,Mangalam_2021_ICCV,Sun_2021_ICCV,Rasouli_2021_ICCV} and vehicles \cite{Phillips_2021_CVPR,Fang_2020_CVPR,Amirloo_2021_CVPR,Kim_2021_CVPR,Cui_2021_ICCV,Dendorfer_2021_ICCV}. 

There are different approaches to designing trajectory prediction models in the literature. Scene-based methods mainly rely on the semantic map of the environment alongside the agents' positional information represented as a point-cloud map \cite{Phillips_2021_CVPR, Cui_2021_ICCV,Casas_2020_ECCV,Ye_2021_CVPR}, or 2D representations of trajectories \cite{Fang_2020_CVPR,Mangalam_2020_ECCV,Mangalam_2021_ICCV}. These approaches often use convolutional networks to encode map and positional information. Point-based approaches, on the other hand, explicitly process agents' positional information as a time-series data using recurrent \cite{Shafiee_2021_CVPR,Sun_2021_ICCV,Sadeghian_2019_CVPR} or transformer-based \cite{Yu_2020_ECCV, Yuan_2021_ICCV,Liu_2021_CVPR} techniques. Unlike scene-based techniques, these methods often explicitly model the interactions between the agents and their surroundings. 

As mentioned earlier, an agent can potentially have equally valid future trajectories, e.g. turning or going straight at an intersection, and it is important to design a  predictive model capable of capturing such diversity. However, this goal is not achievable by  using unimodal approaches \cite{Liang_2019_CVPR,Rasouli_2021_ICCV} or simply constraining the output over a probability distribution \cite{Chandra_2019_CVPR,Mohamed_2020_CVPR}. To remedy this issue,  some algorithms use proposal-based approaches \cite{Fang_2020_CVPR,Narayanan_2021_CVPR, Phan-Minh_2020_CVPR, Pan_2020_IROS,Chai_2019_CORL} in which  predefined trajectory anchors are generated according to the observed dynamics of the agents or map constrains. Although these methods encourage trajectory diversity, they largely rely on heuristic methods and their performance depends on the quality of predefined anchors. 

Another category of algorithms are generative models that learn the distributions over the trajectories. Some of these methods are based on GAN architectures \cite{Dendorfer_2021_ICCV, zhao_2019_CVPR1, Sadeghian_2019_CVPR, Gupta_2018_CVPR} in which adversarial learning is used to estimate distributions of trajectories. However, in addition to the challenges in designing loss functions and training, it is hard to assign semantic meanings to the latent representations in GANs which affects the interpretability of the results \cite{Shen_2020_CVPR}. VAE-based methods \cite{Yuan_2021_ICCV,Bhattacharyya_2021_CVPR,Salzmann_2020_ECCV,Casas_2020_ECCV} address this issue by explicitly learning the latent representations. In practice they are often conditioned on the future states of the agents during the training phase. An alternative approach is using normalizing flow \cite{Ma_2021_ICCV, Park_2020_ECCV} (inspired by \cite{Rezende_2015_ICML}) to estimate distributions over the trajectories. Using this method, the posterior  distributions are constructed using a sequence of invertible transformations. The methods using normalizing flow, however, rely on step-wise sampling which can potentially limit the ability of the network to learn a single ``intention" for each generated sequence. 

Inspired by \cite{Tang_2019_NIPS}, our approach is a conditional variational method that uses past trajectories and map information to form a discrete latent representation. Compared to the previous approaches, we generate one sample at a time $t$  from posterior distribution which leads to learning more interpretable representations of agents' intentions, e.g. turning left/right at an intersection. 

\subsection{Multi-agent interaction modeling}
Interaction modeling is a crucial part of trajectory prediction methods. Classical approaches rely on physics-based methods to model interactions where forces between agents or agents and obstacles determine their paths  \cite{Rudenko_2018_ICRA,Rudenko_2018_IROS,Helbing_1995_Phys}. Recent learning-based approaches rely on alternative techniques. Social pooling \cite{Mangalam_2020_ECCV,Kothari_2021_CVPR,Sun_2020_CVPR_2} is one of the widely used methods in which trajectories of agents within a region are jointly processed to learn the dependencies among them. Graph-based methods \cite{Zheng_2021_ICCV,Sun_2020_CVPR,Li_2020_NeurIPS,Mohamed_2020_CVPR,Kosaraju_2019_NeurIPS} are also widely used where the agents are represented as nodes and connections are established based on information such as relative distances. 

More recently, many methods use attention-based techniques \cite{Yuan_2021_ICCV,Yu_2020_ECCV,Kamra_2020_NeurIPS,Zhang_2019_CVPR,Sadeghian_2019_CVPR} which process the joint representation of agents' past trajectories using, e.g. a self-attention module, to assign importance to the interacting agents. These methods have the added benefit of handling a larger number of agents as shown in \cite{Park_2020_ECCV}.

The majority of the existing approaches only rely on past observations to model interactions and account for future actions implicitly based on the current states of the agents. The lack of such connections between the current and future states of the agents during prediction stage can be problematic since it might result in generating conflicting trajectories for different agents. Therefore it is important to jointly learn the distributions of the trajectories of all the agents to account for such scenarios as shown in \cite{Rhinehart_2019_ICCV}. Our method models both present and future interactions. During inference, at each time step, the predicted locations of all agents in previous steps are fed into a self-attention module to jointly learn the interactions between them. In other words, our method uses a graphical structure connecting the predictions of the agents both spatially and temporally. Concurrent to our work \cite{girgis2021autobots} and \cite{ngiam2021scene} also model agent interactions using transformer. However, they apply separate attention layers to time and agent dimension.

\subsection{Scene Context Encoding}
Encoding scene context is important for trajectory prediction methods. Information such as lane direction, drivable areas or other environmental constraints are key to generate feasible trajectories. In the context of trajectory prediction, scene information can be represented in various formats, such as rasterized maps \cite{Phillips_2021_CVPR,Cui_2021_ICCV,Liang_2020_ECCV_2}, vectorized maps \cite{Liu_2021_CVPR,Gao_2020_CVPR}, instance-level representations \cite{Kim_2021_CVPR}, 2D scene images \cite{Shafiee_2021_CVPR,Dendorfer_2021_ICCV,Mangalam_2021_ICCV} and drivable areas \cite{Park_2020_ECCV}. The maps are often fed into convolutional networks to generate scene representations which in conjunction with other sensory inputs help the method to predict future trajectories. The main drawback of using CNNs to process maps is that they suffer from inductive biases, such as location invariance which is not a desirable feature in a multi-agent prediction setting. Additionally, processing the map separately from state of the agents impacts the ability of the method to effectively reason about agents and scene relationship.

To remedy these issues, recent approaches use attention mechanisms to weight scene features according to the states of the agents \cite{Liu_2021_CVPR, Shafiee_2021_CVPR, Dendorfer_2021_ICCV}. In \cite{Park_2020_ECCV}, the map features are combined with positional encoding and fed into a vision attention module. The generated attention scores are conditioned on the states of the agents which are iteratively updated at every time-step. Our work expands on this idea in two ways. First, we implement a multi-resolution encoding scheme using local and global patches extracted from the map. These patches are then fed into a vision transformer to generate a sequence of learned representations. In order to condition on these representations, the trajectory decoder cross-attends to the learned sequence.

	\section{Problem Formulation}
We assume that the multi-agent system is a partially-observed Markov process and formulate it as a continuous space, discrete time for $A$ variable number of agents. We denote the state of $a$-th agent at time $t$ as $\mathcal{S}_t^a \in \mathbb{R}^{d}$ where $d$ is the dimensionality of each state corresponding to $(x, y)$ coordinates. The agents interact over a prediction horizon of length $T$. We represent the joint distribution of all agents over this time horizon as $\mathcal{S}$ where $\mathcal{S} \doteq \mathcal{S}_{0: T}^{1: A}$. Similarly, we define the past trajectories of agents as $\mathcal{S}_{past}$ where $\mathcal{S}_{past} \doteq \mathcal{S}_{-\tau: 0}^{1: A}$.

\section{Technical Approach}

\begin{figure*}
    \centering
    \includegraphics[width=0.9\textwidth]{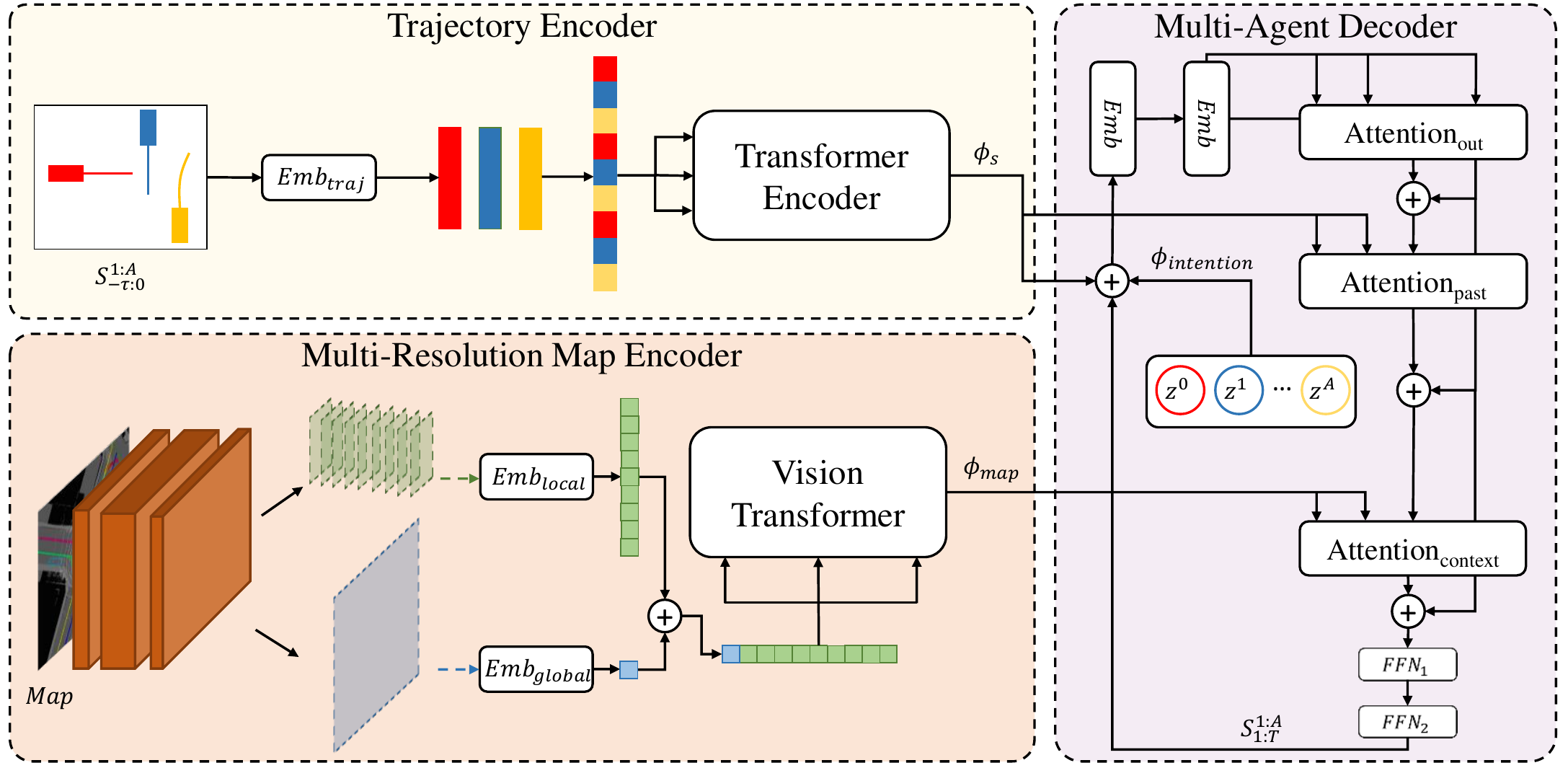}
    \caption{An overview of the proposed model, LatentFormer. There are two modules responsible for encoding the scene context. In the \textbf{trajectory encoder} a mixed embedding of agents' trajectories is fed into a transformer encoder to model the temporal dependencies and interactions. \textbf{Map encoder} processes drivable areas using convolutional layers the output of which is used for global and local representations that are combined and fed into a vision transformer to capture scene context. \textbf{Decoder} module autoregressively uses the encodings in addition to a sampled intention from the latent space to make the final predictions.}
    \vspace{-0.5cm}
    \label{diagram}
\end{figure*}

LatentFormer aims to model a multi-modal distribution over the trajectories of all the agents. As illustrated in  Figure \ref{diagram}, the proposed model consists of three main modules: \textbf{trajectory encoder}, \textbf{multi-resolution map encoder} and \textbf{multi-agent decoder}. These modules are described below.

\subsection{Preliminaries}
LatentFormer is a fully transformer-based model and consequently the building blocks of each module are Multi-head Self-attention $\mathbf{Attn_{MS}}$. Each self-attention head computes features as follows:

\begin{equation}
    \begin{array}{c}
        \operatorname{Attn}(\mathbf{Q}, \mathbf{K},     \mathbf{V})=\operatorname{Softmax}\left(\mathbf{Q K}^{\top}\right)     \mathbf{V}
    \end{array}
    \label{eq:self_att}
\end{equation}

\noindent where $\mathbf{K} \in \mathbb{R}^{n_{k} \times d_{q}}$, $\mathbf{Q} \in \mathbb{R}^{n_{q} \times d_{q}}$ and $\mathbf{V} \in \mathbb{R}^{n_{k} \times d_{v}}$ refer to key, query and value vectors, respectively. Each head, $h_i$ computes the self-attention features using Eq. \ref{eq:self_att}. These features are then concatenated and passed through a linear projection, $\mathbf{W^O}$ to compute the final features.
\begin{equation}
    \begin{gathered}
            \operatorname{Attn_{MS}}(\mathbf{Q}, \mathbf{K}, \mathbf{V})=\text{concat}\left(\text { head }_{1}, \ldots, \text { head }_{h}\right) \mathbf{W}^{O}\\
            \text { head }_{i}=\operatorname{Attn}\left(\mathbf{Q W}_{i}^{Q}, \mathbf{K W}_{i}^{K}, \mathbf{V W}_{i}^{V}\right)
    \end{gathered}
\end{equation}

\noindent where $\mathbf{W}_{i}^{K}$, $\mathbf{W}_{i}^{Q}$ and $\mathbf{W}_{i}^{V}$ are the key, query and values of each head, respectively. With a slight abuse of notation we use  $\mathbf{\operatorname{Attn_{MS}(X)}}$ to denote key, value and query vectors are all a linear projection of the input sequence and $\mathbf{\operatorname{Attn_{MC}(X, X')}}$ where queries are the linear projection of X and key and values are the linear projections of $X'$.

\subsection{Trajectory Encoder}\label{tech-enc}
\begin{figure*}
\centering
\includegraphics[width=0.7\textwidth]{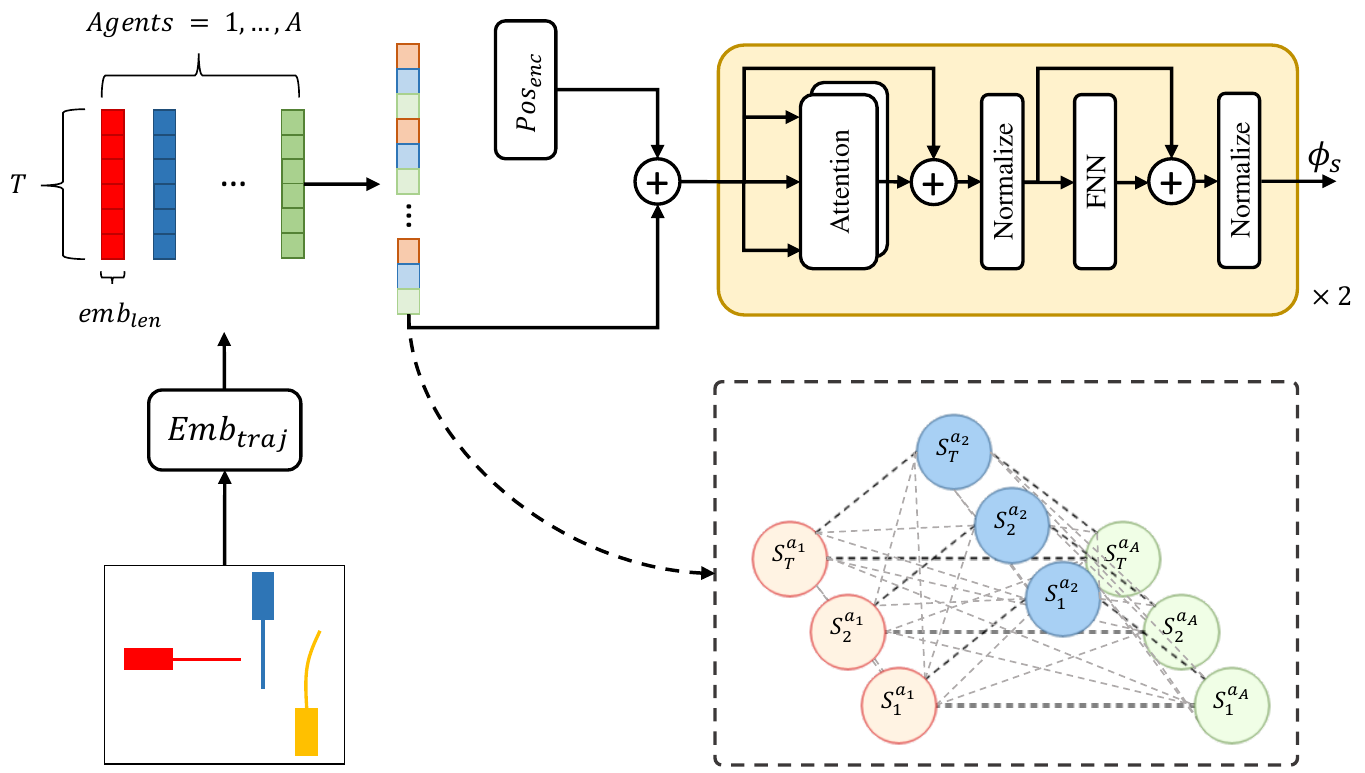}
\caption{An overview of the trajectory encoder module. Agents' trajectories are combined and fed to a linear layer to produce the input embeddings. The learned positional encodings are then added to the input embeddings before being fed to a transformer encoder module. As shown in the figure, the proposed architecture resembles a fully connected graph structure that is capable of capturing complex interactions between the agents. }
\vspace{-0.5cm}
\label{encoder}
\end{figure*}

 The trajectory encoder summarizes the history of trajectories $\mathcal{S}_{past} \doteq \mathcal{S}_{-\tau: 0}^{1: A}$ while modeling the interactions among the agents a transformer encoder. The transformer can be considered as a fully-connected attention-based Graph Neural Network (GNN) where each node is an element in the sequence. Inspired by positional encoding introduced in \cite{vaswani2017attention}, we concatenate the index of each agent to each state, $\mathcal{S}_t^a$, feed it to a linear layer to get the $d_m$ dimension embeddings and restructure the sequence to be a sequence of all agents' trajectories in the observation horizon, as shown in Figure \ref{encoder}. The main advantage of this design is that it can handle a variable number of agents if indexed properly. To ensure that the computational complexity of $n^2$ does not make the system very slow one can limit the number of agents according to the prediction horizon and application. 

The restructured trajectory embedding is added to a learned positional encoding and fed to a transformer module similar to the one described in \cite{vaswani2017attention}. The transformer encoder consists of $i_e$ identical blocks of $\operatorname{TEBlock}$. In the first block, the sequence of encoded inputs are passed to a multi-head self-attention layer and then the result is fed to a position-wise fully-connected feed-forward neural network, $\operatorname{FFN}$. A residual connection connects the input of each layer to its output before a layer normalization \cite{ba2016layer} is applied to it:

\begin{equation}
    \begin{gathered}
    \operatorname{TEBlock}(\mathcal{X}) = \text{LayerNorm}(\mathcal{X}^{\prime} + \operatorname{FFN}(\mathcal{X}^{\prime})) \\
    \mathcal{X}^{\prime} = \text{LayerNorm}(\mathcal{X} + \operatorname{Attn_{MS}}(\mathcal{X}))
    \end{gathered}
\end{equation}
    
\noindent where $\mathcal{X}$ is the input sequence to each block. We denote the output of the encoder as $\phi_{S}$.

\subsection{Multi-Resolution Map Encoder}

\begin{figure}
\centering
\includegraphics[width=0.9\columnwidth]{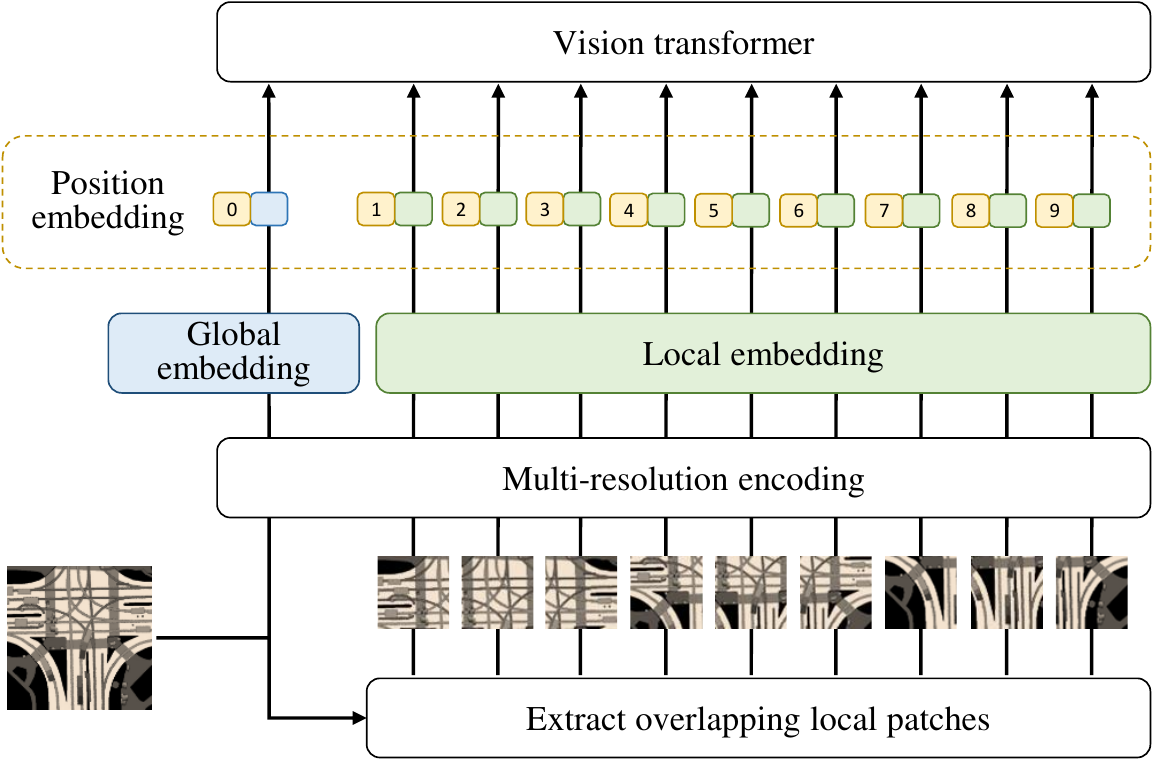}
\caption{An overview of the map encoder module. The map is used at two different levels, a global representation which captures the entire map and local representations which are based on overlapping patches extracted from the map. The combination of both representations is fed into a position embedding layer followed by a vision transformer to generate the final map encoding.}
\vspace{-0.7cm}
\label{map_encoding}
\vspace{-0.1cm}
\end{figure}

As stated earlier, the information encoded in map is extremely valuable, thus using simple semantics like drivable area, if encoded properly, can help the model to avoid unrealistic predictions (e.g. trajectories off the road). Following the approach in \cite{Park_2020_ECCV} we use maps represented as drivable areas in the form of binary masks. The maps are further augmented by adding the absolute pixel indices and the Euclidean distances between each pixel and the center of the map in separate channels. This way the spatial information about each pixel is directly accessible by the network.

The map is first processed using convolutional layers to generate a feature map. This feature map is used to produce local and global embeddings. First, it undergoes a linear transformation to the $d_m$ creating a global embedding $\Gamma_g$ of the whole scene. Next, we leverage an architecture similar to \cite{dosovitskiy2020image} and extract overlapping patches of size $M \times M$ where $M$ is smaller than the feature map size. A linear projection is then applied to these patches to transform them to $d_m$ resulting in the local embedding $\Gamma_l$. A sequence of $\Gamma_g$ and  $\Gamma_l$ is then fed to a vision transformer to generate a sequence of map representations, $\phi_{map}$.

In our experience, this design is more effective than using a single global embedding as each agent's future trajectory may attend to different parts of the map. Additionally, in more complicated datasets one can make a sequence of map embeddings with different resolutions and then let the model learn which resolution is more informative. This eliminates the need for hand designing a cropping size for each agent as in \cite{chai2019multipath}, \cite{Salzmann_2020_ECCV} which is not flexible to different dynamics of the environment, such as speed (e.g. in higher speeds one may choose to use bigger cropping size). 

\subsection{Multi-Agent Decoder}

The goal of prediction is to estimate the joint distribution over future trajectories of all agents conditioned on some context information, in this case observed states ($\phi_{S}$) and map ($\phi_{map}$). To estimate this joint distribution LatentFormer is using $K$ discrete modes for each agent. Each mode is represented as one-hot binary vector. Similar to interaction modeling, the index of each agent is concatenated to each agent's mode representation of size $K$. These vectors are then flattened for all the agents leading to a $A \times K$ length vector with the dimension of $2$. These encodings are then passed to a linear layer and form the intention embeddings $\phi_{intention}$ before getting concatenated with $\phi_s$, $\phi' = concat(\phi_s, \phi_{intention})$. 

LatentFormer computes the distribution over $Z$s conditioned on context, $\phi$, where $\phi$ refers to $\phi_{S}$ and $\phi_{map}$. In order to achieve this, $K$ learnable vectors, $\mathcal{P}$ are utilized for each agent. The index of each agent is also concatenated with this vector and projected to $d_m$ using a linear layer. This sequence is then fed to a transformer decoder block, ${TDBlock_d}$, with two cross-attention components to compute a probability over all $Zs$. 

\begin{equation}
\begin{gathered}
p\left(Z \mid \mathbf{\phi} \right) = \operatorname{Softmax}(\operatorname{FFN}( \operatorname{TDBlock_d}\left(\mathcal{P}, \mathbf{\phi}\right) ))
\end{gathered}
\end{equation}

\noindent where $\operatorname{TDBlock_d}$ consists of $I_{dd}$ identical transformer decoding layers $\operatorname{TDL_{di}}$. Each layer applies the following operations to the input sequence, $X$, where the first input sequence is the linear projection of $\mathcal{P}$:

\vspace{2pt}
\begin{equation}
    \begin{gathered}
    \operatorname{TDL}_{di}(\mathcal{X}, \mathbf{\phi_{S}}, \mathbf{\phi_{map}})= \text{LayerNorm} ( \mathcal{X}' +\mathrm{FF_p}(\mathcal{X}'))) \\
    \mathcal{X}'= \text{LayerNorm}\left(\mathcal{X}'' + \operatorname{Attn}_{MC}\left(\mathbf{\mathcal{X}''}, \mathbf{\phi_{map}} \right)\right) \\
    \mathcal{X}''= \text{LayerNorm}\left(\mathcal{X}'''+ \operatorname{Attn}_{MC}\left( \mathcal{X}''', \mathbf{\phi_{S}}\right)\right) \\
    \mathcal{X}''' = \text{LayerNorm} \left(\operatorname{\operatorname{Attn}_{MS}}(\mathbf{\mathcal{X}})\right)
\end{gathered}
\end{equation}
\vspace{-4pt}

Using this method allows the model to learn different modes of driving in an unsupervised fashion. Intuitively, these modes can represent different driving behaviors and decisions (e.g. going straight, left, right, speed-up, slow-down and etc). In contrast to other works like \cite{Rhinehart_2019_ICCV} and \cite{Park_2020_ECCV} which sample different $z$s at different time steps, in LatentFormer each possible future trajectory for each agent is conditioned on one mode which is more aligned with our goal of encoding intentions of the whole trajectory in the latent space. 

\begin{equation}
\begin{array}{c}
\log p(\mathcal{S} \mid \phi) =
\log \left(\sum\limits_{Z} p(\mathcal{S}, Z \mid \phi\right)= \\
\log \left(\sum\limits_{Z} p(\mathcal{S} \mid Z, \phi) p(Z \mid \phi)\right)
\end{array}
\end{equation}
\vspace{2 pt}

For decoding we use another transformer decoder block which is similar to $TDBlock_c$. However, in our experience, for continuous outputs, layer normalization delays the convergence. Consequently, the second decoder block does not have layer normalization operations. $TDBlock_c$ consists of $I_{dc}$ identical layers, $TDL_{ci}$ with each layer doing the following operations: 

\vspace{2 pt}
\begin{equation}
    \begin{gathered}
    \operatorname{TDL}_{ci}(\mathcal{X}, \mathbf{\phi'}, \mathbf{\phi_{map}})= \mathcal{X}' +\mathrm{FFN}(\mathcal{X}')) \\
    \mathcal{X}'= \mathcal{X}'' + \operatorname{Attn}_{MC}\left(\mathbf{\mathcal{X}''}, \mathbf{\phi_{map}} \right) \\
    \mathcal{X}''= \mathcal{X}'''+ \operatorname{Attn}_{MC}\left( \mathcal{X}''', \mathbf{\phi'}\right)\\
    \mathcal{X}''' = \operatorname{\operatorname{Attn}_{MS}}(\mathbf{\mathcal{X}})
\end{gathered}
\end{equation}
\vspace{2 pt}

Note that the decoder block model follows the same interaction modeling as described for the encoder. Hence, at each time step, the prediction of each agent state is conditioned on all of the other agents. The output sequence is then passed to $N$ feedforward layers to generate the parameters of a bivariate Gaussian distribution at each time step.

\subsection{Learning Objective}

Given the dataset consisting of history of trajectories and context information $\mathcal{O}$, and future trajectories $\mathcal{S}$, $\mathcal{D}=\left\{\left(\mathcal{O}^{(i)}, \mathcal{S}^{(i)}\right), \ldots\right\}_{i=1,2, \ldots,|\mathcal{D}|}$ the trajectory prediction objective is to find the parameters $\theta$ which maximizes the marginal data log-likelihood using maximum likelihood estimation (MLE):

\begin{equation}
\label{eq-original}
\begin{array}{c}
\nabla_{\theta} \mathcal{L}(\theta, \mathcal{D}) \\
=\nabla_{\theta} \log p_{\theta}\left(\mathcal{S} \mid \mathcal{O}\right) \\
=\nabla_{\theta} \log \left(\sum\limits_{Z} p_{\theta}\left(\mathcal{S}, Z \mid \mathcal{O}\right)\right) \\
=\sum\limits_{Z} p_{\theta}\left(Z \mid \mathcal{S}, \mathcal{O}\right) \nabla_{\theta} \log p_{\theta}\left(\mathcal{S}, Z \mid \mathcal{O}\right)
\end{array}
\end{equation}
\vspace{2 pt}

Computing the posterior in Eq. \ref{eq-original}, $p_{\theta}\left(Z \mid \mathcal{S}, \mathcal{O}\right)$, is non-trivial as it varies with $\theta$. Instead $q(Z)$ serves as an approximate posterior. We can then decompose the log-likelihood into the sum of evidence lower bound (ELBO) and the $KL$ distance between the true and estimated posterior \cite{bishop2006pattern}.

\vspace{2 pt}
\begin{equation}
\begin{array}{c}
\log p_{\theta}\left(\mathcal{S} \mid \mathbf{\mathcal{O}} \right)=\sum\limits_{Z} q(Z) \log \frac{p_{\theta}\left(\mathcal{S}, Z \mid \mathbf{\mathcal{O}}\right)}{q(Z)} + \\
D_{K L}\left(q(Z) \| p_{\theta}\left(Z \mid \mathcal{S}, \mathbf{\mathcal{O}}\right)\right)
\end{array}
\end{equation}
\vspace{2 pt}

Similar to \cite{Tang_2019_NIPS} and \cite{girgis2021autobots} we use a form of EM algorithm for optimization. First, we approximate $q(z)$ with $p_{\theta^{\prime}} \left( Z \mid \mathcal{S}, \mathcal{O} \right)$ where $\theta^{\prime}$ are the parameters of the model at the current time step.  Then, we fix the posterior and optimize $\theta$ for the following objective:

\vspace{2 pt}
\begin{equation}
\begin{gathered}
\begin{array}{c}
Q\left(\theta, \theta^{\prime}\right) =\sum\limits_{Z} p\left(Z \mid \mathcal{S}, \mathcal{O} ; \theta^{\prime}\right) \log p(\mathcal{S}, Z \mid \mathcal{O} ; \theta) \\
=\sum\limits_{Z} p\left(Z \mid \mathcal{S}, \mathcal{O} ; \theta^{\prime}\right) (\log p\left(\mathcal{S} \mid Z, \mathcal{O};  \theta\right) +\\ \log p\left( Z \mid \mathcal{O} ; \theta\right));
\end{array}
\end{gathered}
\end{equation}
\vspace{2 pt}

Given that LatentFormer uses discrete latent variable for a given $\theta^{\prime}$, the exact posterior can be computed.

\subsection{Training}
We used teacher forcing to train the model. For teacher forcing, we make a mask which lets the decoder ``see" all the agents in previous time steps. Closer to the convergence we alternate the training to autoregressive mode to make sure the model, when used in evaluation mode, can handle possible mistakes. In addition, we studied the possibility of training the model without any \textit{teacher}. In this scenario, the input sequence to the decoder is not masked and only contains the positional encoding and agents' index embeddings. We showed that this setting takes significantly longer time to converge. However, when it does, it can be used non-autoregressively in inference with a slight decrease in the performance which can be a great advantage in real-world applications.

	\section{Experiments} \label{sec:exp}
In this section, we demonstrate the effectiveness of LatentFormer on real-world data and compare its performance to a number of state-of-the-art approaches. We then investigate the contributions of various aspects of the model on overall performance via ablation studies.

\subsection{Implementation}
LatentFormer includes $2 \times \operatorname{TEBlock}$s, $2 \times \operatorname{TDBlock_d}$ and $4 \times \operatorname{TDBlock_c}$, each with $8$ attention heads and the total embedding dimension of $256$.  The map is processed using $3$ convolutional layers with $8$, $16$ and $6$ filters with the sizes of $3 \times 3$, $3 \times 3$ and $1 \times 1$ and strides of $1$, $2$ and $1$ respectively. The final output of the layers is of dimension $32 \times 32 \times 6$. We set the final embedding dimension to $256$ and the size of mode representation $K$ to 12.
 For training, we used the $SGD$ optimizer with $0.95$ momentum and the initial learning rate of $5e-4$ following the cyclic learning rate schedule \cite{smith2019super}. The batch size was $64$ while training in teacher forcing mode and $16$ in the autoregressive mode.

\subsection{Dataset}
We evaluate the performance of LatentFormer on the nuScenes dataset \cite{Caesar_2020_CVPR}. This dataset consists of 850 driving sequences 20 seconds each. The data contains 3D bounding box annotations for all dynamics objects as well as semantic maps of the environments. Overall, there are a total of 25K trajectory samples. Following the evaluation protocols in the recent works \cite{Park_2020_ECCV}, we extract sequences of $5s$ each comprised of $2s$ observation and $3s$ prediction with the sampling rate of $2Hz$.

\subsection{Metrics}
We report the distance-based metrics widely used in the literature for evaluation. These metrics include Average Distance Error ($\mathrm{ADE}$) and Final Distance Error ($\mathrm{FDE}$) which compute the Euclidean distance between the predicted trajectories and the ground truth. For both metrics we report the results as the average ($\mathrm{avg}$) or minimum ($\mathrm{min}$) calculated over $K$ trajectory samples. Following \cite{Park_2020_ECCV}, we also report the results using $\mathrm{RF}$ metric defined as $\mathrm{avgFDE/minFDE}$. This metric can differentiate between prediction with multiple modes and a single mode.

\subsection{Methods}
Given the multi-modal nature of the proposed method, we compare the performance of our model to the following state-of-the-art approaches:\textbf{ Deep Stochastic IOC RNN Encoder-decoder framework (DESIRE)} \cite{Lee_2017_CVPR}, \textbf{Multi-Agent Tensor Fusion (MATF)} \cite{zhao_2019_CVPR1}, \textbf{Prediction Conditioned on Goals (PRECOG)} \cite{Rhinehart_2019_ICCV}, and \textbf{Diverse and Admissible Trajectory Forecasting (DATF)} \cite{Park_2020_ECCV}. We refer to our proposed model as \textbf{LatentFormer} throughout the text. 

\subsection{Multi-modal Trajectory Prediction}

\begin{table}
	\footnotesize
	\begin{center}
		\begin{tabular}{cc||ccc}
			\multicolumn{2}{c||}{Methods}&  minADE & minFDE & RF  \\
			\hline
			\multicolumn{2}{c||}{DESIRE \cite{Lee_2017_CVPR}} & 
			1.08 & 1.84 & 1.72  \\
			\hline
			\multicolumn{2}{c||}{MATF \cite{zhao_2019_CVPR1} } & 
			1.05& 2.13 & 1.19   \\
			\hline
			\multicolumn{2}{c||}{PRECOG \cite{Rhinehart_2018_ECCV}} & 
			1.18& 2.19 & 1.63   \\
			\hline
			\multicolumn{2}{c||}{DATF \cite{Park_2020_ECCV}} & 
			0.64& 1.17 & \textbf{2.56} \\
			\hline
			\multicolumn{2}{c||}{LatentFormer (Ours)} & \textbf{0.38}& \textbf{0.96}& 2.51
	\end{tabular}
		\vspace{-0.5cm}
	\end{center}
	\caption{The evaluation of LatentFormer on the nuScenes dataset. Lower values on metrics \textit{minADE} and \textit{minFDE} are better and higher values are better on \textit{RF}.}
	\vspace{-0.5cm}
	\label{tab:sota}
\end{table}

We begin by comparing the performance of the proposed method against state-of-the-art algorithms on the nuScenes dataset. The results of the experiment are summarized in Table \ref{tab:sota} and show that LatentFormer achieves state-of-the-art performance on both $\mathrm{minADE}$ and $\mathrm{minFDE}$ by improving the results by up to $40\%$ and $18\%$ respectively while achieving similar performance to DATF on diversity metric, $\mathrm{RF}$.

Such improvements can be attributed to two main features of LatentFormer. Compared to all the models (except for DATF), in addition to the global map, our model also locally reasons about different regions of the map allowing the method to better capture the scene structure. Moreover, LatentFormer jointly predicts future states of the agents resulting on generating non-conflicting trajectories which is not the case in the DATF model.  

\subsection{Ablation Study}

\begin{figure*}
\centering
\includegraphics[width=0.9\textwidth]{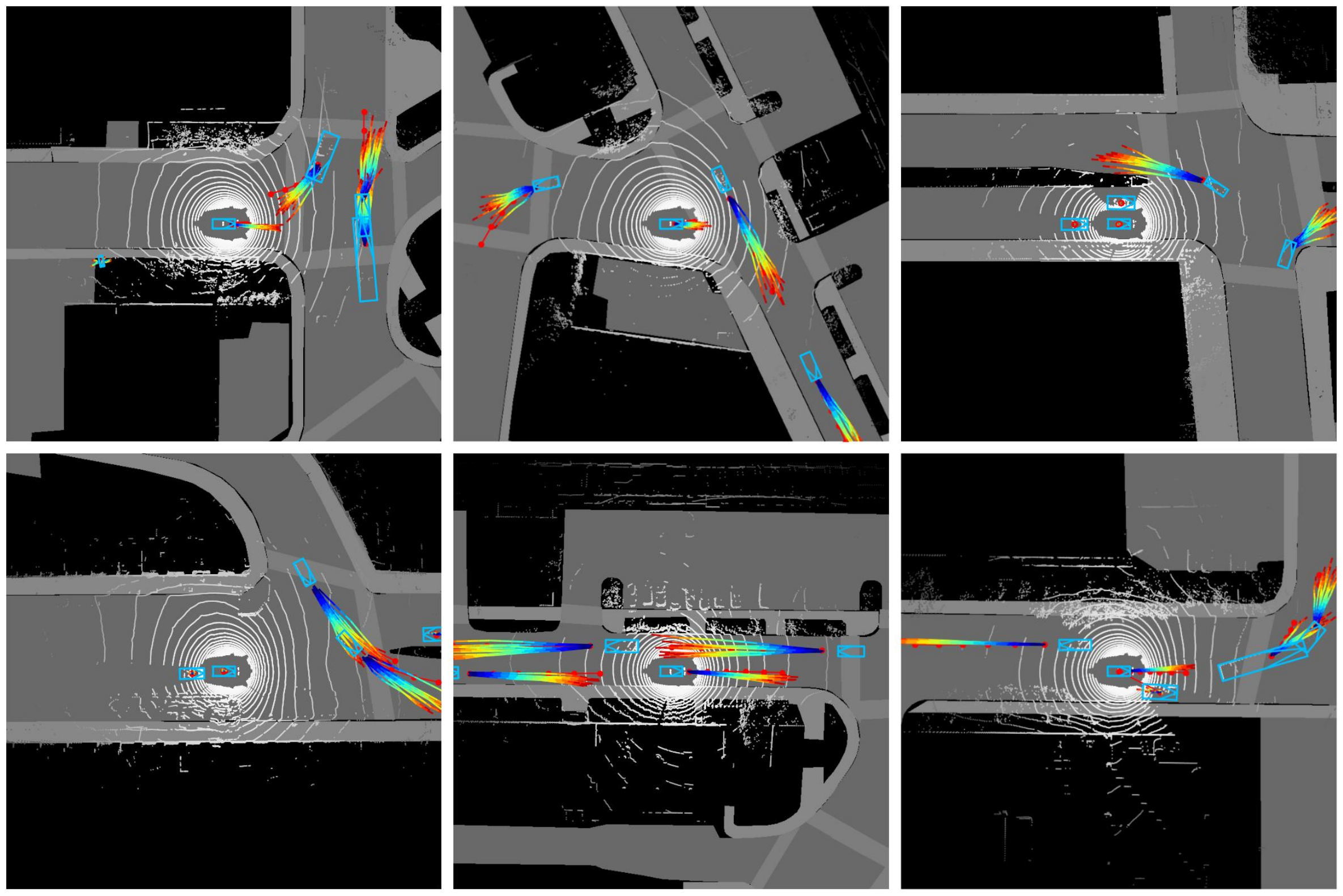}
\caption{The qualitative results of the LatentFormer on nuScenes. The maps are centered on the ego-vehicle and all vehicles are shown using blue rectangles. Triangles within rectangles show the orientation of the corresponding agents. The colors of trajectories reflect the predictions at different time steps (from blue for earlier to red for later). The ground truth trajectories are shown in solid red color.}
\vspace{-0.3cm}
\label{qual_res}
\end{figure*}

\begin{table*}[!t]
	\footnotesize
	\begin{center}
		\begin{tabular}{cc||cccc||cccc}
			\multicolumn{2}{c||}{Architectures}& \multicolumn{2}{c}{TransMap} & \multicolumn{2}{c||}{Interaction}& minADE & avgADE & minFDE & avgFDE \\
			\hline
			\multicolumn{2}{c||}{LatentFromer-encdec} & 
            \multicolumn{2}{c}{\xmark} &
            \multicolumn{2}{c||}{\xmark}&
			0.83 & 1.80 & 1.37 & 3.33 \\
			\hline
            \multicolumn{2}{c||}{LatentFromer-CNNMap} & 
            \multicolumn{2}{c}{\xmark} &
            \multicolumn{2}{c||}{\xmark}&
			0.75 & 1.59 & 1.23 & 3.00 \\
			\hline
			\multicolumn{2}{c||}{LatentFromer-TransMap} & 
            \multicolumn{2}{c}{\cmark} &
            \multicolumn{2}{c||}{\xmark}&
			0.52 & 1.07 & 1.09 & 2.16 \\
			\hline
			\multicolumn{2}{c||}{LatentFromer-Interaction} & 
            \multicolumn{2}{c}{\xmark} &
            \multicolumn{2}{c||}{\cmark}&
			0.76 & 1.26 & 1.27& 2.56 \\
			\hline
			\multicolumn{2}{c||}{LatentFromer} & 
            \multicolumn{2}{c}{\cmark} &
            \multicolumn{2}{c||}{\cmark}&
			\textbf{0.38} & \textbf{0.96} & \textbf{0.72} & \textbf{1.81} \\
		\end{tabular}
			\vspace{-0.5cm}
	\end{center}
	\vspace{-5pt}
	\caption{Evaluation of LatentFormer with different proposed components for map encoding and interaction modeling. On all metrics, lower value is better. }
	\label{tab:abl-cm}
	\vspace{-10pt}
\end{table*}

\textbf{Map encoding.} In this study, we compare the performance of three alternative approaches for map encoding. As a baseline, we use the model without any map information and we refer to it as \textit{LatentFormer-encdec}. The second approach, \textit{LatentFormer-CNNMap},  is based on the common technique in existing methods in which only a global map of the environment $\Gamma_g$ is encoded using CNNs. We refer to our multi-resolution encoding with vision transformer method as \textit{LatentFormer-TransMap}.

As shown in Table \ref{tab:abl-cm}, using map in any format is advantageous. Here we can see that LatentFormer-CNNMap outperforms the baseline model LatentFormer-encdec which has no map information. However, due to the limitations of CNNs, such as inductive bias, this technique does not suffice to model the interactions of the agents with the environment. Using local map representation along with a positional embedding and a vision transformer remedy this issue. As a result, LatentFormer-TransMap achieves significantly better performance improving all metrics by up to $30\%$. 


\textbf{Interaction Modeling} One way to evaluate the effectiveness of the interaction modeling is to compare the performance of the algorithm when each agent's trajectory is predicted independently versus when the trajectories of all the agents are jointly predicted. For this experiment we add interaction modeling to the LatentFormer-encdec and keep everything else the same. We refer to this model as LatentFormer-Interaction. Table \ref{tab:abl-cm} shows that LatentFormer works better when it explicitly models interactions.

\begin{table}[!t]
    \label{abl-aut}
	\footnotesize
	\begin{center}
		\begin{tabular}{cc||cccc}
			\multicolumn{2}{c||}{Architectures} &  minADE & avgADE & minFDE & avgFDE \\
			\hline
			\multicolumn{2}{c||}{AutoRegressive} &  0.38 & 0.96 & 0.72 & 1.81 \\
			\hline
			\multicolumn{2}{c||}{Non-autoregressive} &  0.42 & 1.01 & 1.05 & 2.00 \\
			\hline
		\end{tabular}
			\vspace{-0.4cm}
	\end{center}
	\vspace{-5pt}
	\caption{Evaluation of LatentFormer using alternative decoding methods.}
	\label{tab:abl-auto}
	\vspace{-10pt}
\end{table}

\textbf{Autoregressive vs non-autoregressive.} We studied an alternative architecture of our model. In this version the whole sequences of the states are predicted simultaneously. This lets the model to condition on both future and past embeddings of all the states while allowing parallel computation. The results are summarized in Table \ref{tab:abl-auto}. During training, the autoregressive model has converged after $40$ epochs whereas non-autoregressive model took $90$ epochs to be fully trained. Although the performance of the latter is slightly worse and training can become more challenging on more complicated datasets,  the possibility of parallel computation in real-time may make non-autoregressive model a better candidate for real-world applications.

\subsection{Qualitative Results}

A robust predictive model should be capable of generating feasible trajectories under different conditions. Figure \ref{qual_res} illustrates how the proposed method achieves this goal. In this figure, we show the output of our model in different scenarios, e.g. driving straight or turning. To better highlight how the predictions progress over-time, we used a color spectrum from blue to red. 

As expected, the generated trajectories for each agent diverge as the predictions reach longer time horizon. The majority of these predictions, however, are admissible as the proposed method can effectively reason about the map structure and the states of the agents to make predictions.

	\section{Conclusion}

In this paper, we proposed LatentFormer, a probabilistic framework in which we estimate the joint distribution over future trajectories of all agents in a multi-agent setting. LatentFormer leverages the flexibility and the power of attention based architectures to model interactions and use the map information while decoding sequence of states. A possible avenue for the future work is to to make the model more computationally efficient in the scenes with a larger number of agents and longer prediction horizon by a hierarchical design where each group of interacting agents are identified in initial steps.

{\small
\bibliographystyle{ieee_fullname.bst}
\bibliography{latentformer.bib}
}

\end{document}